\documentclass[letterpaper, 10 pt, conference]{ieeeconf}  

\IEEEoverridecommandlockouts                              

\usepackage{amsmath} 
\usepackage{amssymb}  
\usepackage{todonotes}
\usepackage{algorithm} 
\usepackage{algpseudocode} 
\usepackage{hyperref}
\usepackage{soul}

\title{\LARGE \bf
Dynamic Inference on Graphs using Structured Transition Models
}

\author{Saumya Saxena$^{1}$ and Oliver Kroemer$^{1}$
\thanks{*This work was in part supported by the National Science Foundation under Grant No. CMMI-1925130. Any opinions, findings, and conclusions or recommendations expressed in this material are those of the author(s) and do not necessarily reflect the views of the NSF.}
\thanks{$^{1}$Robotics Institute, School of Computer Science, Carnegie Mellon University, Pittsburgh PA 15123, USA {\tt\small \{saumyas, okroemer\}@andrew.cmu.edu}}%
}

\begin{document}

\maketitle
\thispagestyle{empty}
\pagestyle{empty}

\begin{abstract}

Enabling robots to perform complex dynamic tasks such as picking up an object in one sweeping motion or pushing off a wall to quickly turn a corner is a challenging problem. The dynamic interactions implicit in these tasks are critical towards the successful execution of such tasks. Graph neural networks (GNNs) provide a principled way of learning the dynamics of interactive systems but can suffer from scaling issues as the number of interactions increases. Furthermore, the problem of using learned GNN-based models for optimal control is insufficiently explored. In this work, we present a method for efficiently learning the dynamics of interacting systems by simultaneously learning a \textit{dynamic} graph structure and a \textit{stable} and \textit{locally linear} forward model of the system. The dynamic graph structure encodes evolving contact modes along a trajectory by making probabilistic predictions over the edges of the graph. Additionally, we introduce a temporal dependence in the learned graph structure which allows us to incorporate contact measurement updates during execution thus enabling more accurate forward predictions. The learned stable and locally linear dynamics enable the use of optimal control algorithms such as iLQR for long-horizon planning and control for complex interactive tasks. Through experiments in simulation and in the real world, we evaluate the performance of our method by using the learned interaction dynamics for control and demonstrate generalization to more objects and interactions not seen during training. We introduce a control scheme that takes advantage of contact measurement updates and hence is robust to prediction inaccuracies during execution.



\end{abstract}
\section{INTRODUCTION}

Common everyday tasks such as picking up an object in one smooth motion, pushing a heavy door using the momentum of our bodies, or pushing off a wall to quickly turn a corner involve complex dynamic interactions between the human and the environment. These dynamic interactions are critical in successful execution of these tasks. Thus, to enable robots to perform such dynamic tasks effectively, we need to consider the dynamics of the robot, the individual objects, as well as the interactions between them. One way to accomplish this is to explicitly learn these complex dynamics from data, and then use the learned models to intentionally leverage contacts to perform interactive and dynamic tasks.

Learning the interaction dynamics of multi-body systems is a challenging problem especially as the number of interacting bodies in the scene increases, resulting in a combinatorial explosion in the size of the state space as well as in the number of interactions to reason about. In recent work, graph neural networks (GNNs) \cite{battaglia2016interaction, battaglia2018relational} have been used to model and learn \textit{object-centric} interaction dynamics of such systems. These networks provide a principled way of learning the dynamics of large interactive systems by incorporating a strong structural prior (\textit{relational inductive bias}) in the learned dynamics. However, many GNN based approaches still suffer from scaling issues as the number of interactions increases \cite{hoshen2017vain}. For real-world dynamic manipulation tasks which usually involve a small set of interactions occurring at a given time in the scene, learning time varying sparse graph structures can help mitigate some of these issues.

In this work, we develop a method for efficiently learning the dynamics of interacting systems by simultaneously learning a relational \textit{dynamic} graph structure and a \textit{stable locally linear} forward dynamic model of the system. The learned dynamic graph structure encodes information about dynamics and contact modes evolving along the task trajectory by adding and removing edges from the graph as contact is made and broken. This approach allows for strong generalization of the learned dynamics when more nodes are added to the graph. Starting by representing our system as a fully connected graph, we learn time-varying and state dependent \textit{edge activations} using a \textit{graph inference module}. An edge is inferred to be active when the nodes connected via the edge interact and influence the forward dynamics, thus making it critical for the edge activations and the forward model to be learned together. When a new object is added to the scene, the graph inference module can predict whether or not the new node influences the system dynamics and the dynamics of other nodes based on edge predictions. This allows the learned model to generalize more effectively to new objects and new interactions in the environment.

We use the spring-mass-damper model as a structural prior over the \textit{local} object-centric dynamics of our system. This provides us with two benefits: linearity and stability. For positive values of mass, stiffness and damping parameters, the linear spring-mass-damper model is always stable about the equilibrium point (which is also learned) \cite{aastrom2010feedback}. Assuming a locally linear and stable prior structure over our dynamics model allows us to represent a wide range of non-linear systems and use the learned dynamics for long-horizon planning and control by precluding the rollout trajectories from growing unbounded. The learned locally linear dynamics also enable the use of optimal control algorithms, such as iLQR, to exploit the interaction dynamics and perform complex tasks optimally by predicting future edge activations and thus contact modes as well.

Control strategies that use a learned model are naturally sensitive to model inaccuracies, especially near the contact regions. This can lead to catastrophic failures during execution. To alleviate this issue, we use gated recurrent units (GRUs) to learn a probabilistic model that predicts edge activations at each time step. These units introduce a temporal dependency of the current edge activation on the previous edge activations. During execution, observed contact modes are used to evaluate a posterior over the edge activations and update them, which enables more accurate future rollouts. We use model predictive control (MPC) to replan using the observed current state and contact modes.

We also explore an exciting application of this work in the field of apprenticeship learning wherein the learned locally linear model can be used for learning the quadratic cost function underlying expert demonstrations using differentiable LQR as the policy class \cite{amos2018differentiable, saxena2021learning}. The learned behavior can then be generalized to unseen goal conditions.

The key contributions of this work are two-fold: 1) a method for learning stable locally linear dynamics for non-linear interactive systems using graph neural networks by encoding changing dynamics and contacts as part of the graph structure, enabling strong generalization properties to more objects in the scene, and 2) using the learned locally linear dynamics to devise a robust control scheme that utilizes the recurrent nature of the learned graph structure to adapt the model predictions and the policy to observed contact events.

We evaluate our results in simulation on multiple object dynamic pickup and dynamic door-opening tasks. We also perform real world experiments using the 7DOF Franka-Emika Panda robot arm for a dynamic pickup task  \footnote{Supplementary video: https://youtu.be/A9YG8VrIpgQ}.
\section{RELATED WORK}
Learning the interaction dynamics of contact-rich tasks is a problem of interest in many areas of robotics research such as manipulation \cite{kroemer2019review}, collaborative robotics and assistive systems \cite{dragan2017robot}. In some recent works the dynamics of such interactive systems is learned implicitly \cite{fragkiadaki2015learning, lerer2016learning}, while in others \cite{saxena2021learning, khader2020data, toussaint2005learning,kroemer2015towards} the focus is on modelling these changing dynamics as explicit dynamic modes. In our method, we aim to exploit the inherent structure underlying interactions in physical systems by modelling them as graphs and taking advantage of the generalization benefits of graph-based approaches.

There has been tremendous progress towards the development of graph neural networks for modelling and learning the forward interaction dynamics of physical systems. Approaches such as \cite{battaglia2016interaction, sanchez2020learning, battaglia2018relational, weng2021graph, watters2017visual} learn the dynamics model assuming a static fully connected or known graph structure. Such models can require a large amount of interaction data and training time to accurately learn the interaction dynamics. There has also been work towards using static graphs for planning and control \cite{sanchez2018graph, wang2018nervenet, janner2018reasoning}. We focus on the problem of simultaneously learning the dynamic graph structure and the forward dynamic model of the system in a purely unsupervised fashion and using that model for control.

Some approaches \cite{li2020causal, kipf2018neural, alet2019neural} learn a graph structure that is static over the entire task, or along a single task trajectory, while other methods learn to actively predict edge interactions \cite{hoshen2017vain, van2018relational, goyal2019recurrent} using attention mechanisms. We take inspiration from these approaches for learning probabilistic predictions of edge activations that evolve with the state of the system.

Learning to control non-linear systems using locally optimal control algorithms, such as iLQR, is an exciting area of research. Such methods require learning locally linear and stable dynamics models of the system. Some recent approaches that aim to learn locally linear dynamics of the system for control include \cite{watter2015embed,banijamali2018robust, zhang2019solar, levine2014learning} but these approaches have rarely been extended to graph based interactive dynamic systems. We derive inspiration from these works and model our system dynamics as locally linear, which enables the use of optimal control algorithms such as iLQR. In our work, we also explore methods developed in \cite{amos2018differentiable, saxena2021learning, yin2021policy} for utilizing the learned linearized dynamics for learning simple quadratic cost functions of differentiable controllers such as LQR.

\begin{figure*}
    \centering
    \includegraphics[width=\textwidth]{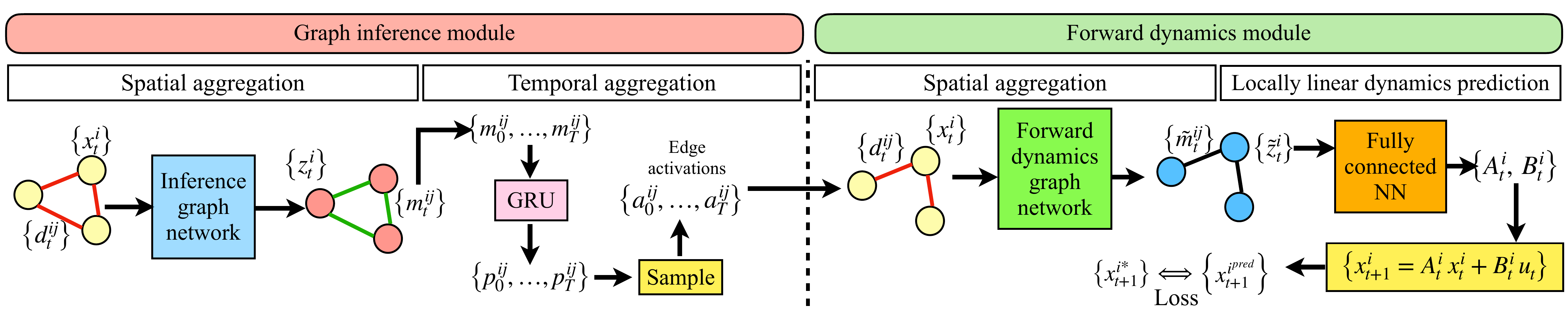}
    \caption{Method overview: Our method has two main parts 1) Graph inference module: we start with a set of fully connected graphs and pass them through the inference graph network that performs message passing between the nodes and outputs embedded graphs. Next, to aggregate information temporally, we pass the edge embeddings $\mathbf{m}_{0,\dots,T}^{ij}$ through a GRU that outputs a discrete probability distribution $\mathbf{p}_t^{ij}$ over the edge types (active or inactive) for each edge. We sample from this distribution to get the edge activations $\mathbf{a}_t^{ij}$ (see \ref{sec:graphinf}). 2) Forward dynamics module: using the edge activations, we remove the inactive edges from the fully connected input graphs and then pass them through the forward dynamics graph network that outputs another set of embedded graphs. The node embeddings $\tilde{\mathbf{z}}_t^{ij}$ are passed though a fully connected neural network that outputs stable locally linear transition dynamics (see \ref{sec:fwddyn}). We forward propagate the learned dynamics using the input control and get the next state which is used for loss calculation during training \eqref{eq:loss}.}
    \label{fig:method}
\end{figure*}
\section{METHOD}
In this section, we discuss in detail our method for learning stable locally linear dynamics for non-linear interactive systems using graph neural networks. Our training pipeline is composed of two main modules that are trained together, 1) a graph inference module that learns the dynamic graph structure, and 2) a forward dynamics module that learns the stable locally linear forward dynamics given the graph structure. 

\subsection{Graph Inference Module}
\label{sec:graphinf}
This module predicts the dynamic interaction graph structure --- which edges in the dynamic graph are \textit{active} or \textit{inactive} at a certain time step. In our domain, where physical objects are represented as nodes in the graph, this module infers whether or not, at a certain time step, objects influence each other's dynamics. For example, in a scenario where a robot gripper picks up an object, an edge connects the gripper and the object only after the object has been grasped.

The state of the system at time $t$ is represented as a graph $\mathcal{G}_t = (\mathcal{N}_t, \mathcal{E}_t)$ where nodes $\mathcal{N}_t$ represent objects in the scene and edges $\mathcal{E}_t$ connect interacting nodes. Node $n^i_t \in \mathcal{N}_t$ is represented using a feature vector $\mathbf{x}_t^i$ where $\mathbf{x}_t^i = [\mathbf{q}_t^i, \mathbf{\dot{q}}_t^i]$. $\mathbf{q}_t^i$ and $\mathbf{\dot{q}}_t^i$ are the position and velocity of the $i$th object respectively. The edge $e_t^{ij} \in \mathcal{E}_t$ connecting nodes \textit{i} and \textit{j} is represented with edge features $\mathbf{d}_t^{ij}$ where $\mathbf{d}_t^{ij}$ is the distance between nodes $n^i_t$ and $n^j_t$.

An overview of the graph inference module is shown in Fig. \ref{fig:method}. We start with a set of trajectories of fully connected graphs $(\mathcal{G}_0, \mathbf{u}_0, \dots, \mathcal{G}_T, \mathbf{u}_T)$ where $\mathbf{u}_t$ is the control applied at time t, and $T$ is the length of the trajectory.
In order to learn the dynamic graph structure, we first use a graph neural network to propagate information spatially over each of the graphs $(\mathcal{G}_0, \dots, \mathcal{G}_T)$ via message passing between the nodes using a node feature network $f_n^{\text{inf}}$ and an edge feature network $f_e^{\text{inf}}$:
\begin{equation*}
    \begin{aligned}
    \mathbf{m}_t^{ij} &= f_e^{\text{inf}}(\mathbf{x}_t^i,\mathbf{x}_t^j,\mathbf{d}_t^{ij}) \\
    \mathbf{z}^i_t &= f_n^{\text{inf}}(\mathbf{x}_t^i, \sum_j{\mathbf{m}_t^{ij}})
    \end{aligned}
\end{equation*}
where $f_e^{\text{inf}}$ and $f_n^{\text{inf}}$ are fully connected neural networks representing the \textit{Inference GNN} and $\mathbf{z}^i_t$ and $\mathbf{m}_t^{ij}$ are the output node and edge embeddings respectively. After two steps of message passing, the output edge embeddings $\mathbf{m}_{0:T}^{ij}$ for each edge are passed though a gated recurrent unit (GRU) to allow temporal information flow:
\begin{equation}
\begin{aligned}
\label{eq:gru}
\mathbf{p}_{0:T}^{ij} &= \text{GRU}_{\phi}(\mathbf{m}_{0:T}^{ij})\\
\mathbf{a}_{0:T}^{ij} &= \text{sample}(\mathbf{p}_{0:T}^{ij})
\end{aligned}
\end{equation}
where $\phi$ are the trainable parameters of the GRU and $\mathbf{p}_{t}^{ij}$ is a discrete probability distribution over the edge types. We consider two edge types: active and inactive. Sampling from probability distribution $\mathbf{p}_{t}^{ij}$ gives us a one hot vector representing the \textit{edge activation} $\mathbf{a}_{t}^{ij}$. If an edge is active at time t, $\mathbf{a}_{t}^{ij}=[1, 0]$ else $\mathbf{a}_{t}^{ij}=[0, 1]$. Naturally, this representation can be extended to include more than two edge types. Differentiability of this sampling procedure can be ensured by employing techniques such as Gumbel softmax \cite{jang2016categorical}. If an edge $e_t^{ij}$ is sampled to be inactive it is removed from the graph $\mathcal{G}_t$ for the subsequent calculations and no messages pass through that edge. The output of the graph inference module is a trajectory of graphs $(\tilde{\mathcal{G}}_0, \dots, \tilde{\mathcal{G}}_T)$ where $\tilde{\mathcal{G}}_t = (\mathcal{N}_t, \mathcal{\tilde{E}}_t)$ and $\mathcal{\tilde{E}}_t \subseteq \mathcal{E}_t$ is the set of active edges. Given this inferred graph structure, our next step is to model the forward dynamics of the system. 

\subsection{Forward Dynamics Module}
\label{sec:fwddyn}
An overview of the forward dynamics module is shown in Fig. \ref{fig:method}. To learn the object-centric interaction dynamics of the system, we use another GNN to perform two steps of message passing over each of the graphs $(\tilde{\mathcal{G}}_0, \dots, \tilde{\mathcal{G}}_T)$ to propagate information spatially between connected nodes.
\begin{equation*}
    \begin{aligned}
    \tilde{\mathbf{m}}_t^{ij} &= f_e^{\text{dyn}}(\mathbf{x}_t^i,\mathbf{x}_t^j,\mathbf{d}_t^{ij}) \\
    \tilde{\mathbf{z}}^i_t &= f_n^{\text{dyn}}(\mathbf{x}_t^i, \sum_j{\tilde{\mathbf{m}}_t^{ij}})
    \end{aligned}
\end{equation*}
where $f_e^{\text{dyn}}$ and $f_n^{\text{dyn}}$ are fully connected neural networks representing the \textit{Forward Dynamics GNN}. Here, for scenarios where more than two edge types are considered, for example to capture varied behavior over active edges, one can use a different edge feature network $f_e^{\text{dyn}}$ for each edge type. The output node embeddings $\tilde{\mathbf{z}}^i_t$ now encode interaction information between the connected nodes and can be used to learn the forward dynamic parameters of the interactive system independently for each node. To learn stable locally linear dynamics, we approximate the local dynamics of each node to resemble a spring-mass-damper system that is decoupled in its degrees of freedom. The parameters for each degree of freedom are derived from the node embeddings at each time step. This representation was chosen since the dynamics of a spring-mass-damper system can be constrained to be stable (bounded output) by constraining the mass, stiffness and damping parameters to be greater than zero \cite{aastrom2010feedback}. This is easily achieved by using activations such as sigmoid and relu. We use a fully connected neural network $g^{\text{dyn}}$ to output the dynamic parameters for each node as,
\begin{equation}
\label{eq:params_out}
\boldsymbol{\alpha}^i_t = g^{\text{dyn}}(\tilde{\mathbf{z}}^i_t)
\end{equation}
where $\boldsymbol{\alpha}^i_t = [\frac{1}{m^i_t}, k^i_t, c^i_t, \hat{x}_{t}^i]$ where $m^i_t$ is the mass, $k^i_t$ is the stiffness of the spring, $c^i_t$ is the damping, $\hat{x}_t^i$ is the equilibrium point for node $i$ at time $t$. These parameters can then be used to compute the stable locally linear dynamics of the system as:
\begin{equation}
    \begin{aligned}
    \mathbf{x}^i_{t+1} &= \mathbf{A}^i_{t} \mathbf{x}^i_{t} + \mathbf{B}^i_{t} \mathbf{u}_{t} + \mathbf{o}^i_{t}
    \end{aligned}
    \label{eq:fwddyn}
\end{equation}
where \begin{equation*}
    \mathbf{A}^i_{t} = \begin{bmatrix}
    1 & dt\\
    \frac{-k^i_t \, dt}{m^i_t} & 1 - \frac{c^i_t \, dt}{m^i_t}\\
    \end{bmatrix},
    \mathbf{B}^i_{t} = \begin{bmatrix}
    0\\
    \frac{dt}{m^i_t}\\
    \end{bmatrix},
    \mathbf{o}^i_{t} = \begin{bmatrix}
    0\\
    \frac{k^i_t \, \hat{x}_{t}^i \, dt}{m^i_t}\\
    \end{bmatrix}
\end{equation*}
for a discrete-time one degree of freedom system.

\subsection{Training}
The graph inference module and the forward dynamics module are trained together to simultaneously learn the dynamic graph structure and the locally linear forward dynamics. We collect a set of $N$ trajectories containing multiple interacting objects and represent them as graphs $\{(\mathcal{G}_0^*, \mathbf{u}_0, \dots, \mathcal{G}_{T-1}^*, \mathbf{u}_{T-1})_n\}_{n=0,\dots, N}$. The loss function for training the model is written as, 
\begin{equation}
    \label{eq:loss}
    \mathcal{L}_\text{dyn} = \sum_i ||\mathbf{x}^{i^{*}}_{1:T} - \mathbf{x}^{i^{\text{pred}}}_{1:T}||_2 + \sum_{i, j} \text{KL}[\mathbf{p}_{0:T-1}^{ij} \, || \, \mathbf{q}_{0:T-1}^{ij}]
\end{equation}
where $\mathbf{x}^{i^*}_{t+1}$ is the observed next state and $\mathbf{x}^{i^{\text{pred}}}_{t+1} = \mathbf{A}^i_{t} \mathbf{x}^{i^{*}}_{t} + \mathbf{B}^i_{t} \mathbf{u}_{t} + \mathbf{o}^i_{t}$ is the predicted next state for node $n^i$. $\mathbf{p}_{0:T-1}^{ij}$ is the probability distribution over predicted edge activations and $\mathbf{q}_{0:T-1}^{ij}$ is a prior on the edge activations. In prior work \cite{li2020causal, kipf2018neural}, where the focus is on learning a static graph structure, the prior is designed such that a sparse graph is learned. However, in our scenario, since the graph structure is not static and can be sparse or dense depending upon the state, we design a prior using the relative distances between the nodes. In particular, when two nodes are spatially close to each other, the prior suggests that the probability of the edge connecting these nodes being active, is high. In particular, for a given distance threshold of $d_{th}$, the prior over edge $e^{ij}_t$ at time t is given by: 
\begin{equation*}
    \mathbf{q}_{t}^{ij} = \text{Softmax}(
    \begin{bmatrix}
    d_{th} - d_t^{ij}\\
    d_t^{ij} - d_{th}
    \end{bmatrix}
    )
\end{equation*}
where the edge feature $d_t^{ij}$ is the distance between nodes $n^i_t$ and $n^j_t$. The neural networks  $f_e^{\text{inf}}$, $f_n^{\text{inf}}$, $\text{GRU}_{\phi}$, $f_e^{\text{dyn}}$, $f_n^{\text{dyn}}$ and $g^{\text{dyn}}$ are trained simultaneously using stochastic gradient descent to minimize the loss $\mathcal{L}_\text{dyn}$. 

\subsection{Execution}
\label{sec:exec}
While it is common practice to use model predictive control for replanning using the current observed state, for interactive tasks where contacts are critical, it is essential to utilize observed contact information for replanning as well. Since we use GRU to learn the edge activations, it introduces a temporal dependency of the current edge activation on the previous edge activations. For one time step prediction, this recurrent process looks as follows, $\mathbf{p}_{t+1}^{ij} = \text{GRUCell}(\mathbf{p}_{t}^{ij}, \mathbf{m}_{t}^{ij})$. During execution, using the observed contact modes, we update the current probabilistic edge activations by calculating a posterior on the edge activations as follows,
\begin{equation*}
\label{eq:posterior}
    \tilde{\mathbf{p}}_{t}^{ij} = z \, \mathbf{p}_{t}^{ij} \, \mathbf{c}_t^{ij}
\end{equation*}
where $\mathbf{c}_t^{ij}$ is a discrete probability distribution representing the observed contact between node $n_i$ and $n_j$ at time $t$ and $z$ is the normalization constant.  Since GRU aggregates past information for future rollouts, the updated edge activation $\tilde{\mathbf{p}}_{t}^{ij}$ allows for more accurate future predictions. In particular, we use the posterior edge activations $\tilde{\mathbf{p}}_{t}^{ij}$ to make updated future predictions as follows, $\mathbf{p}_{t+1}^{ij} = \text{GRUCell}(\tilde{\mathbf{p}}_{t}^{ij} , \mathbf{m}_{t}^{ij})$. 

\begin{figure*}[t]
    \centering
    \includegraphics[width=\textwidth]{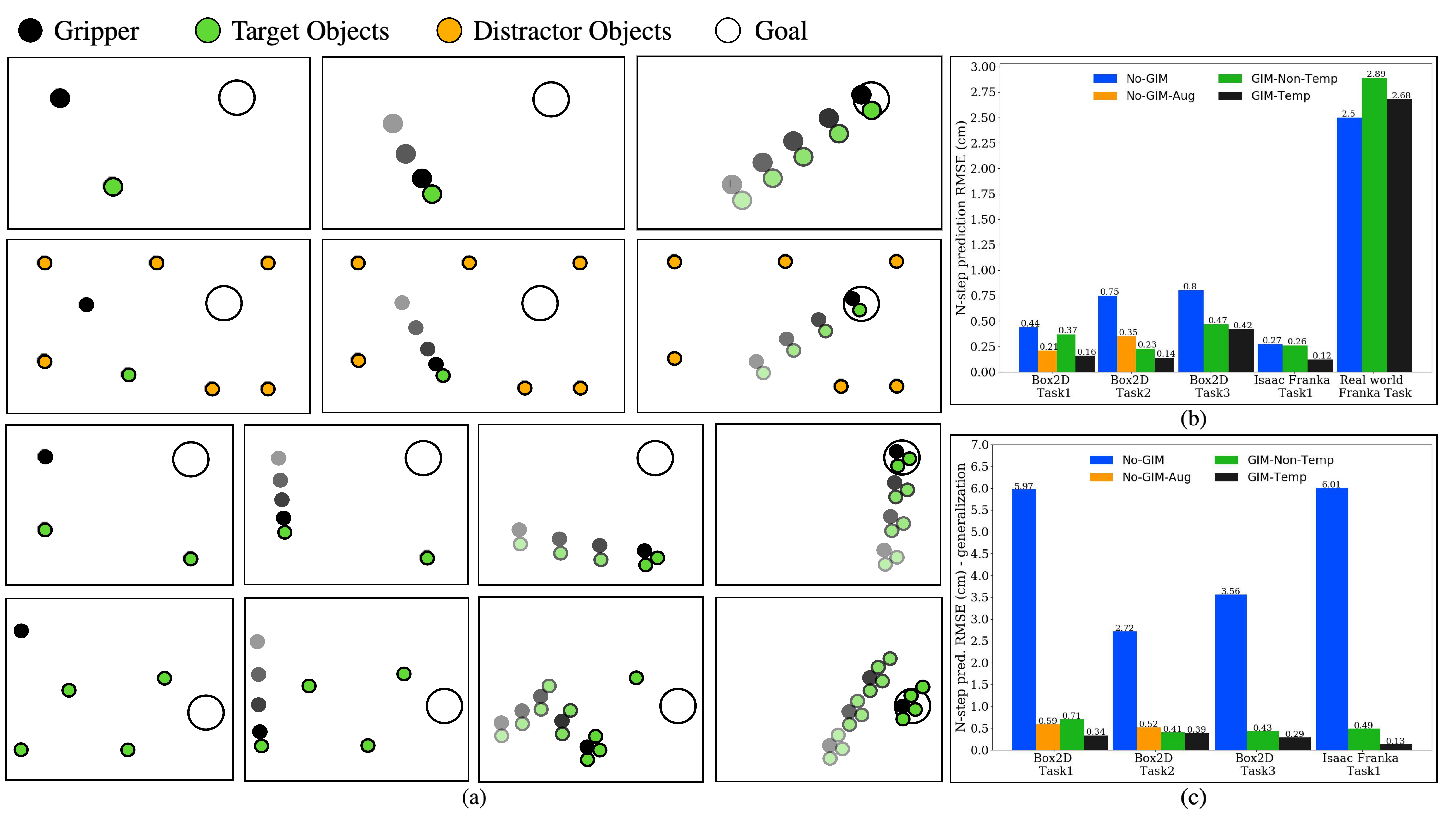}
    \caption{(a) Row 1: Box2D Task 1 with same number of objects as seen during training. Row 2: Box2D Task 1 generalization scenario with many distractor objects (orange) in the scene. Row 3: Box2D Task 2 with same number of objects as seen during training. Row 4: Box2D Task 2 generalization scenario for picking up many more objects than during training. (b) N-step root mean square error between the predicted and executed trajectories. These experiments are performed with the same number of objects and interactions as seen during training. (c) N-step root mean square error between the predicted and executed trajectories. These experiments are performed to test for generalization by including many more objects and interactions in the task than seen during training.}
    \label{fig:box2d_and_error}
\end{figure*}
\section{EXPERIMENTS}
\label{sec:experiments}
The proposed method is evaluated in simulation using Box2D and Isaac Gym environments and in the real-world on the 7DOF Franka-Emika Panda robot arm. Through our experiments, we aim to evaluate 1) How useful is our graph inference module in enabling efficient and accurate learning of the dynamics of the interactive system? 2) What are the generalization properties of the learned dynamics model as more objects (not seen during training) are added to the system? 3) How robust is our execution strategy to inaccuracies in the learned model?  4) Can our learned stable linearized dynamics be used for learning the cost function of a differentiable LQR controller using expert demonstrations of a task?

\textbf{Ablation studies:} To answer the first three questions we perform the following three ablation studies:
1) Training the dynamic model without the graph inference module i.e. learning the forward dynamics module given fully connected graphs. The dataset for this study is the same as the dataset used for training the full model. We call this model \textbf{No-GIM}. 2) Training the dynamic model without the graph inference module but with an augmented dataset containing more objects and interactions. We call this model \textbf{No-GIM-Aug}. \textbf{No-GIM} and \textbf{No-GIM-Aug} are similar to prior work \cite{battaglia2016interaction, sanchez2020learning, sanchez2018graph} that assume a static fully connected or known graph structure. 3) Training the full dynamic model with the graph inference module but replacing the GRU in \eqref{eq:gru} with a fully connected neural network that takes as input only the current edge embedding $\mathbf{m}_t^{ij}$ to output the edge activation i.e. there is no temporal relationship between the predicted edge activations. We call this model \textbf{GIM-Non-Temp}. Due to this modification in the network architecture, we will no longer be able to utilize the robust control scheme elaborated on in \ref{sec:exec} since posterior on the edge activations can no longer be used to update future predictions. We call our model \textbf{GIM-Temp}. In addition to the above ablation studies, we also train a model \textbf{Full-AB} that uses our graph inference module but instead of using a spring-mass-damper system based stable dynamics model in the forward dynamics module, learns the full $\mathbf{A}$ and $\mathbf{B}$ matrices from scratch. 

\textbf{Application: Apprenticeship learning.} To answer question 4, given the learned stable locally linear interaction dynamics, we learn the cost matrices $\mathbf{Q}$ and $\mathbf{R}$ of a discrete-time linear quadratic regulator (LQR) using expert demonstrations. We demonstrate generalization of learned expert behavior to goal regions not visited by the expert. We closely follow the cost learning method in \cite{saxena2021learning}, but instead of learning a different cost function and goal condition for each dynamic \emph{mode}, modeled as a discrete global variable, we learn a single cost function for a certain task with a fixed goal.

\textbf{Evaluation metrics:} To test for generalization, all the ablation models mentioned above, including our model, are tested with more objects and interactions than seen during training. For evaluating the accuracy of the learned model, the generalization properties and the effectiveness of the robust control scheme, we use N-step prediction error as the metric. This error is calculated upon using the learned locally linear models for control using iLQR-MPC (receding horizon control) and calculating the root mean square error between the predicted trajectory and the executed trajectory for a rollout length of $N$. For each environment, $N$ is calculated such that the rollout time is 0.5 seconds i.e $N=0.5/dt$. The prediction error is calculated over all the nodes in the graph. For \textbf{GIM-Non-Temp}, in addition to N-step prediction error, we also evaluate the accuracy of the predicted edge activations for a rollout length of $N$ and compare it with our method \textbf{GIM-Temp}. We get the ground truth edge activations by observing contacts during execution. Contact information is not used during training. For \textbf{No-GIM-Aug}, we also present results for the sample complexity, i.e how much more interaction data is needed in order to achieve similar level of performance as with our method. This will demonstrate the efficiency of our method as compared to models where the graph inference module is not used.

\begin{figure*}
    \centering
    \includegraphics[width=0.9\textwidth]{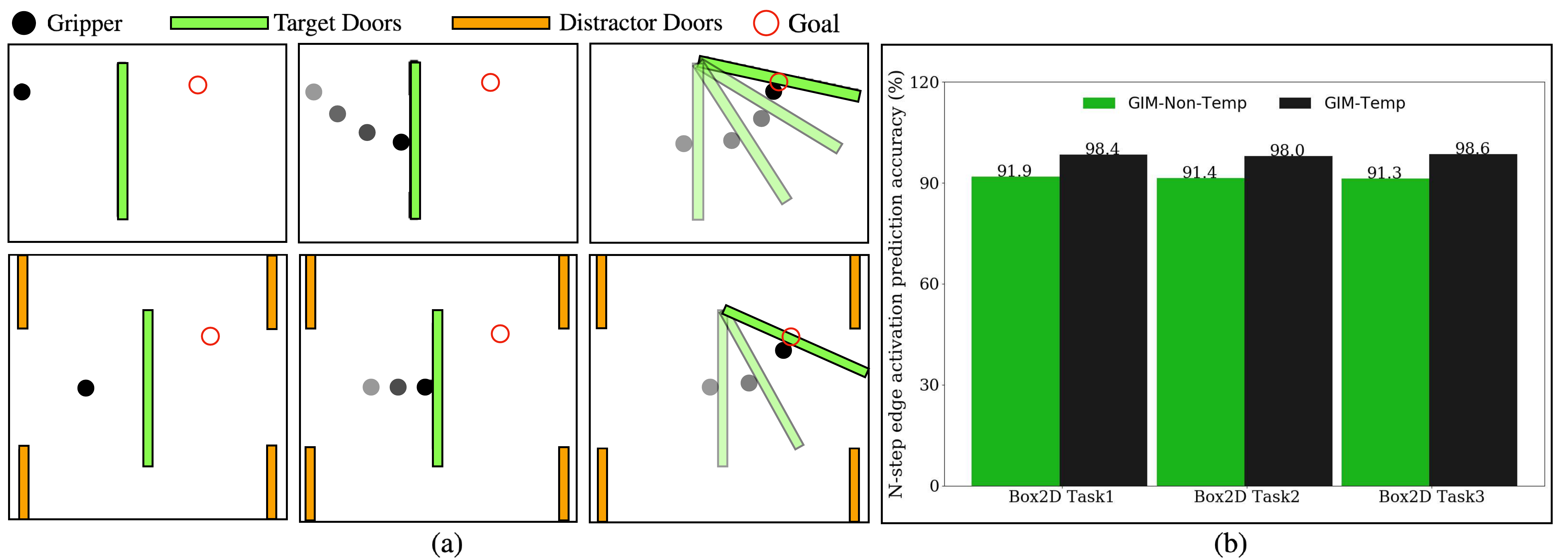}
    \caption{(a) Row 1: Box2D Task 3 with same number of doors as seen during training. Row 2: Box2D Task 3 generalization scenario with many distractor doors in the scene. (b) N-step edge activation prediction accuracy.}
    \label{fig:doors_and_accuracy}
\end{figure*}

Fig. \ref{fig:box2d_and_error}(b) shows the N-step prediction error when the learned models are tested in environments with the same number of objects and interactions as seen during training. Fig. \ref{fig:box2d_and_error}(c) shows the N-step prediction error for generalization scenarios i.e. testing with more interactions and more distractor objects in the scene. Fig. \ref{fig:doors_and_accuracy}(b) shows the N-step edge activation prediction accuracy.

\textbf{Box2D Tasks:} We consider three tasks in this environment: \textbf{Task 1:} Dynamically (without stopping) picking up a 2D object using a 2D gripper and taking it to a goal region (Fig. \ref{fig:box2d_and_error}(a) Rows 1 \& 2). This task aims to assess if the learned graph inference module generalizes well to non-interacting objects in the scene (no edges to non-interacting objects) by testing with up to ten distractor objects. For this task, \textbf{No-GIM}, \textbf{GIM-Non-Temp} and \textbf{GIM-Temp} are trained on a dataset consisting of 40 trajectories of the gripper picking up \textit{one} target objects and taking it to goal location. Initial positions of the gripper and object and goal locations are sampled to collect data. \textbf{No-GIM-Aug} is trained with an augmented dataset consisting of up to three distractor objects which do not interact (160 trajectories). \textbf{Task 2:} Dynamic pickup of multiple objects (Fig. \ref{fig:box2d_and_error}(a) Rows 3 \& 4). This task is designed to test for generalization to more \textit{interactions} than seen during training. We test this by picking up upto five objects during testing while only two objects were picked up in the training dataset. This generalization is non-trivial, as compared to task 1 where distractor objects are ignored, since the model needs to have learned the dynamics behind the interactions that occur when an object is picked up and then generalize it to additional such interactions in the trajectory. For this task, \textbf{No-GIM}, \textbf{GIM-Non-Temp} and \textbf{GIM-Temp} are trained on a dataset consisting of 40 trajectories of the gripper picking up \textit{two} target objects and taking them to a goal. \textbf{No-GIM-Aug} is trained with an augmented dataset consisting of trajectories where the gripper picks up a maximum of three objects and takes them to a goal (120 trajectories). \textbf{Task 3:} Dynamic door opening (Fig. \ref{fig:doors_and_accuracy}(a)). This task aims to access the model's ability to learn locally linear dynamics of a highly non-linear system and use the learned dynamics for control. We add upto ten distractor objects in the test setup of this task as well. For this task, \textbf{No-GIM}, \textbf{GIM-Non-Temp} and \textbf{GIM-Temp} are trained on a dataset consisting of 40 trajectories of the gripper dynamically opening a door.

\textbf{Isaac Franka Task:} We consider the task of dynamically picking up an object of mass 0.5kg using a 7DOF Franka-Emika Panda arm and taking it to a goal location (Fig. \ref{fig:franka_tasks}(a) and (b)). The robot is controlled using task space impedance control \cite{zhang2020modular} during data collection wherein the orientation of the gripper is kept fixed. The initial cartesian position of the gripper, object (on a table) and goal are sampled for data collection. \textbf{No-GIM}, \textbf{GIM-Non-Temp} and \textbf{GIM-Temp} are trained on a dataset consisting of 200 trajectories of the gripper picking up the target object and taking it to a goal. A cartesian space model of the robot is learned while assuming fixed orientation. We test for generalization in environments with up to five distractor objects.

\textbf{Real robot experiments:} Real world experiments are performed for the task of dynamically picking up an object of mass 0.83kg and taking it to a goal location using a 7DOF Franka-Emika Panda robot arm (Fig. \ref{fig:franka_tasks}(c)). The setup, controller, data collection and training strategy is the same as for the Isaac Gym Franka pickup task. The position and velocity of the object were tracked using April tags and a Kinect depth sensor \cite{wang2016apriltag}. Due to lack of reliable contact sensing, we do not update edge activation predictions using observed contacts during execution but still rely solely on GRU for edge activation predictions. \textbf{No-GIM}, \textbf{GIM-Non-Temp} and \textbf{GIM-Temp} are trained on a dataset of 25 trajectories.

\textbf{Training parameters and network architecture:} All the models are trained using ADAM \cite{kingma2014adam} with default parameters and a learning rate of $1e-5$. The inference GNN and forwards dynamics GNN are composed of two fully connected neural networks each having two layers with 64 units each and ReLU activation. $g^{\text{dyn}}$ in \eqref{eq:fwddyn} is also a fully connected neural network having two layers with 64 units each and ReLU activation. We use \textit{GRUCell} in PyTorch \cite{paszke2019pytorch} and apply a softmax to its output for learning the edge activation probabilistic model. 


\section{RESULTS AND DISCUSSION}

For each of the tasks, we use the learned dynamic models for control using iLQR-MPC and calculate the N-step prediction error as shown in Fig. \ref{fig:box2d_and_error}(b \& c). We observe that, when tested with the same number of object interactions as seen during training, the N-step prediction error for all the models trained using simulation data is less than a centimeter and around 2.5cm for the real world Franka task (Fig. \ref{fig:box2d_and_error}(b)). However, when tested for generalization, the performance of \textbf{No-GIM} deteriorates significantly across all the tasks as shown in Fig. \ref{fig:box2d_and_error}(c). For the Isaac Gym Franka task, control using the learned model is not able to successfully pick up the object in many trials when distractor objects are present in the scene. This is because in the absence of the graph inference module, \textbf{No-GIM} cannot infer that edges to additional objects that do not interact can be rendered inactive. This model also fails to infer the implicit mechanism behind the dynamic interaction that occurs when an edge becomes active and hence cannot generalize to additional objects. We also observe that \textbf{No-GIM-Aug} generalizes well and that \textbf{No-GIM-Aug} and \textbf{GIM-Non-Temp} have comparative performance. Training with augmented data consisting of multiple interactions and distractor objects helps \textbf{No-GIM-Aug} ignore distractor objects and generalize to more interactions, however, this comes at the cost of three to four times more data and longer training time. In comparison, \textbf{GIM-Non-Temp} and our model \textbf{GIM-Temp}, trained on the same dataset as \textbf{No-GIM} generalize well to scenarios involving more interactions and distractor objects than seen during training. 
\begin{figure*}
    \centering
    \includegraphics[width=0.9\textwidth]{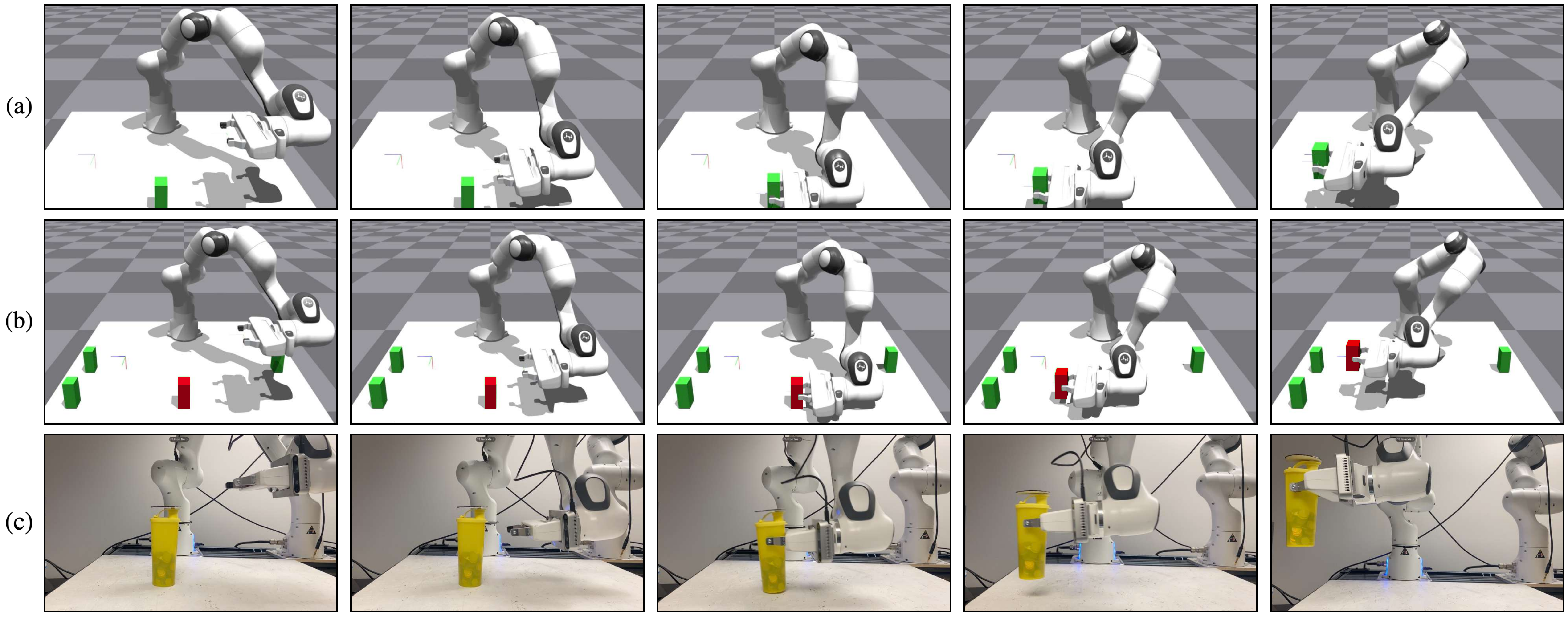}
    \caption{(a) Isaac Gym Task with same number of objects as seen during training. (b) Isaac Gym Task generalization scenario with distractor (green) objects in the scene. (c) Real world experiments with the 7DOF Franka-Emika Panda arm performing a dynamic pickup task using our learned model.}
    \label{fig:franka_tasks}
\end{figure*}

Overall, the N-step prediction performance of \textbf{GIM-Temp} exceeds all the other models in all tasks except for the real world Franka task. Since we do not use contact information for updating the edge predictions for the real world Franka task, the performance of our method, compared to other ablation studies, is not as good as in the other tasks. However, using iLQR-MPC with our learned linear model, we were able to perform the pickup task successfully as shown in Fig. \ref{fig:franka_tasks}(c). This demonstrates the benefit of learning a dynamic graph structure that encodes temporal dependence in the edge activations and of using a robust execution strategy that adapts to observed contact events. This can be quantitatively observed in Fig. \ref{fig:doors_and_accuracy}(b). Both \textbf{GIM-Non-Temp} and our model \textbf{GIM-Temp} show interpretable results where edges were learned to be active only between objects and agents in contact. However, the edge activation prediction accuracy of our model is higher as compared to \textbf{GIM-Non-Temp}. This is because \textbf{GIM-Non-Temp} shows inaccuracies in regions close to contact and, in the absence of a mechanism for updating future predictions based on contact observations, its performance suffers.

For the \textbf{Full-AB} model, trained using our graph inference module but instead of using a spring-mass-damper system based stable dynamics model learns the full $\mathbf{A}$ and $\mathbf{B}$ matrices from scratch, we observed that even though the model trained to convergence, using this learned model for LQR was infeasible. This is because, without any constraints or prior structure in $\mathbf{A}$ and $\mathbf{B}$ ensuring stability, the learned model causes the predicted trajectories to become unstable, resulting in unbounded behavior upon forward rollout.

We also trained a model similar to our model \textbf{GIM-Temp} and using the same data but without including the KL divergence loss between the predicted edge activations and the distance-based prior in the loss function \eqref{eq:loss}. We observe that without an informative prior, the model tends to quickly fall into a local minima where the graph stays fully connected for all time steps. This results in a model learned similar to \textbf{No-GIM} with poor generalization performance.

Finally, we use our learned model \textbf{GIM-Temp} to learn the quadratic cost matrices of a LQR controller using the method described in \cite{saxena2021learning} for the Box2D pickup task 1. We learn the cost function underlying this dynamic task using expert demonstrations collected by sampling the initial position of the gripper but keeping the goal and object locations fixed. Our aim is to evaluate if the learned controller mimics the behavior of the expert and if it generalizes to unseen goal and object locations. We evaluate the performance of the learned controller using task success as the metric. If the gripper is able to pick up the object and take it within a goal region in time T then the task is considered successful. We observe that the learned controller can perform the task with 98\% success rate.
\section{LIMITATIONS AND FUTURE WORK}
One of the limitations of our method is that it cannot, in the present state, incorporate high-dimensional sensor inputs such as images. Similar to prior work \cite{watter2015embed,banijamali2018robust, zhang2019solar, levine2014learning}, we plan to extend our method to end-to-end learn a low-dimensional feature vector from sensory inputs along with the dynamic graph structure and the forward dynamic model. Using long-horizon predictions to reason about occluded or invisible objects is another exciting direction for future work \cite{watters2017visual}. The dependence of our control strategy on reliable contact measurement is another direction of improvement, since reliable contact measurements are not always available. One direction that can be considered is using the approach in \cite{sugimoto2012emosaic} to evaluate the probability of seeing the observed transition for all possible edge activations and use that to calculate the posterior. Another exciting future direction is learning a graph network based LQR controller that encodes relative costs between connected nodes and a goal node and implicitly infers the combined cost over tasks involving varying number of objects.

\section{CONCLUSION}

In this work, we demonstrated the benefits of learning the forward model of interactive systems by simultaneously learning a dynamic interaction graph and a stable locally linear forward dynamic model given the graph. We showed that using a spring-mass-damper system as an approximation to the local dynamics allows us to effectively learn the parameters of a stable system that can be used for long-horizon planning. We then elaborated on the generalization benefits of learning a dynamic graph structure over using a fully connected graph and demonstrated that by introducing a temporal dependency of the current edge activation on the previous edge activations, we allow for contact measurement updates, ensuring more accurate future predictions. As an application of our work, we also discussed how our learned model can be used to learn the quadratic cost function underlying expert demonstrations using the method described in \cite{amos2018differentiable, saxena2021learning}. The learned behavior can then be generalized to unseen goals.


\section*{ACKNOWLEDGMENT}
The authors would like to thank Kevin Zhang and Blake Buchanan for their support during this work.

\bibliographystyle{IEEEtran}
\bibliography{references}

\newpage
\section*{SUPPLEMENTARY MATERIAL}

\section*{EXPERIMENT DETAILS}

\textbf{Box 2D data:} The trajectory length for all the box2D tasks is 10 seconds with $dt=0.05$. The simulation environment is setup such that when two objects come in contact they undergo a purely inelastic collision and a rigid link is formed between them, thus simulating grasping. We collect 40 trajectories for training and 10 for validation by sampling the initial position of the objects/doors, the gripper and the goal in a 2m $\times$ 2m 2D space. The masses for the gripper and objects are 1 kg each. The batch size used for training is 64.

\textbf{Issac Gym data:} The trajectory length for the Isaac Gym Franka task is 4 seconds with $dt=0.01$. We collect 200 trajectories for training and 20 for validation by sampling the initial position of the objects and the goal in a 0.4m $\times$ 0.7m $\times$ 0.3m 3D region of interest in front of the robot. The mass of the object for pickup is 0.4 kg. The batch size used for training is 256.

\textbf{Real Franka data:} The trajectory length for the Real world Franka task is 10 seconds with $dt=0.001$. We collect 25 trajectories for training and 5 for validation by sampling the initial position of the objects and the goal in a 0.4m $\times$ 0.7m $\times$ 0.3m 3D region of interest in front of the robot. The mass of the object for pickup is 0.83 kg. The batch size used for training is 128.

\section*{ADDITIONAL RESULTS}
\subsection*{Learned dynamic parameters and edge predictions}
For the Box2D tasks 1 \& 2, the dynamics are linear and represent that of a point-mass system. We visualize the learned mass, stiffness and damping parameters and the predicted edge activations for a trajectory in Fig. \ref{fig:mode}. The gripper reaches the object and makes contact at around time step 50 as observed in the position plots as well as in the edge activation plot (top left in Fig. \ref{fig:mode}). The edge between the gripper and the object is inactive before contact and active after contact. Before contact the mass of the gripper is predicted to be 1kg and the object mass of predicted to be 0kg. This is because since we apply the same control to all the nodes of the graph, actuated or unactuated, the object mass needs to be zero such that the effect of the actions on it is zero before contact. At the moment of contact we observe an impulse in the velocity plots wherein an inelastic collision happens between the gripper and the object. This is also reflected as a spike in the dynamic parameter plots and represents impact parameters. After contact and hence pickup, the masses of the gripper and the object are predicted to be 0.5kg each. This is because since the control effort is now applied to a total mass of 2kg, each node is effectively sharing half the control effort, which reflects as their individual masses being halved. The other parameters $k_x, k_y, c_x$ and $c_y$ are predicted to be zero (except at impact) as expected for a point mass system.

\section*{CONTROL USING LEARNED LOCALLY LINEAR MODELS}
As mentioned in \ref{sec:experiments} each of the learned models are used for control using iLQR-MPC. The iLQR procedure is as follows:

\begin{algorithm}
\caption{Iterative LQR}
\begin{algorithmic}
\State \scalebox{1.0}{$(i)$ represents the iteration number}
\State \scalebox{1.0}{T is the length of the trajectory}
\State \textbf{Given}: Initial state $\mathbf{x}_{init}$.
\State \textbf{Parameters}: \scalebox{1.0}{$\mathbf{A}_t, \mathbf{B}_t,  \mathbf{x}_f, \mathbf{Q}_t, \mathbf{R}_t $}
    
\Procedure{iLQR}{Parameters}
    \While{not converged}
        \State \!\!\!\!\scalebox{1.0}{$\mathbf{x}^{(i)}_{0:T},\mathbf{K}_{0:T-1}$ $\gets$ LQR($\mathbf{A}_t, \mathbf{B}_t,  \mathbf{x}_f, \mathbf{Q}_t, \mathbf{R}_t$)}
        \State \!\!\!\!\scalebox{0.83}{converged $\gets$ $||\mathbf{x}_{0:T}^{(i+1)} - \mathbf{x}_{0:T}^{(i)}||_2 + ||\mathbf{u}_{0:T-1}^{(i+1)} - \mathbf{u}_{0:T-1}^{(i)}||_2$ is small}
    \EndWhile
\EndProcedure
\end{algorithmic}
\label{alg:ilqr}
\end{algorithm}

The MPC procedure is given in Algorithm 2. Note that the observed contact modes are used to evaluate a posterior over the edge activations and this posterior is used for future edge activation predictions.

\begin{algorithm}
\caption{Model Predictive Control}
\begin{algorithmic}
\State $T_{LQR}$ is time horizon for execution before replanning
\State $N_{MPC} = T/T_{LQR}$
\State $T$ is the length of the trajectory
\State $t$ is the current time step

\State $t \gets 0$
\State \textbf{Given}: Initial state $\mathbf{x}_{init}$, Initial edge activation $\mathbf{p}_0^{ij}$, Learned model $\mathbf{T}$
\State \textbf{Parameters}: \scalebox{1.0}{$\mathbf{x}_f, \mathbf{Q}_t, \mathbf{R}_t $}
\For{n in $N_{mpc}$}
    \State \scalebox{1.0}{$\mathbf{A}_t, \mathbf{B}_t  \gets \mathbf{T}(\mathbf{p}_t^{ij}, \mathbf{x}_{t})$}
    \State \scalebox{1.0}{$\mathbf{K}_{t:T}$ $\gets$ iLQR($\mathbf{x}_t,\mathbf{A}_t, \!\mathbf{B}_t\!,
    \! \mathbf{R}_t,
    \!\mathbf{Q}_t,
    \!\mathbf{x}_f$)}
    \For{i in $T_{LQR}$}
        \State $\mathbf{a}_t \gets \mathbf{K}_t(\mathbf{x}_t - \mathbf{x}_f)$
        \State $\mathbf{x}_{t+1} \gets \text{step}(\mathbf{a}_t)$
        \State $\mathbf{\tilde{p}}_t^{ij}\gets z \mathbf{p}_t^{ij},\mathbf{c}_t^{ij}$
         \Comment{ Posterior \eqref{eq:posterior}}
        \State $\mathbf{p}_{t+1}^{ij} = \text{GRUCell}(\mathbf{\tilde{p}}_t^{ij}, \mathbf{d}_t^{ij})$
        \State $t \gets t_{t+1}$
    \EndFor
\EndFor
\end{algorithmic}
\label{alg:mpc}
\end{algorithm}

\section*{COST LEARNING}

In this section, we talk about how we utilize the learned stable linearized dynamics of the system, to learn the quadratic cost matrices of LQR using expert demonstrations. We closely follow the cost learning method in \cite{saxena2021learning}, but instead of learning a different cost function and goal condition for each dynamic \emph{mode}, modeled as a single discrete global variable, we learn a single cost function for a certain task with a fixed goal. Similar to \cite{saxena2021learning}, we aim to learn the \textit{behavior} of the expert in a single scenario and generalize that behavior to new goal conditions. 

We collect a set of 40 expert demonstrations $\{\mathcal{G}_0^*, \mathbf{u}_0^*, \dots, \mathcal{G}_T^*, \mathbf{u}^*_T\}$. Using the learned graph inference module, we first evaluate the edge activations $\{a_{0}^{ij},\dots,a_{T}^{ij}\}$ over the expert trajectories. We check for the isolated nodes i.e. nodes that are not actuated and have no active edges connected to them in any of the trajectories at any time step and remove those nodes from all the expert graphs. This is based on the indication that these isolated nodes are not participating in the current task. Then, we append together the node attributes of participating nodes at each time step to get the full state of the system as a vector. These full state trajectories are then used as input expert demonstrations to the method described in \cite{saxena2021learning} for learning the quadratic cost matrices $Q$ and $R$. Since the input to LQR is the full appended state of the system, the diagonal terms in the learned $Q$ and $R$ matrices will correspond to penalties for being away from the goal and the non-diagonal terms will correspond to penalties for being away from other nodes. Thus, for tasks like picking up an object with an arm, the learned $Q$ and $R$ enable relative motion between the nodes and then motion towards the goal.

\begin{equation}
\begin{aligned}
& \underset{\mathbf{u}_t}{\text{min}}
& & \!\!\!\!\sum_{t=0}^T \big( (\mathbf{x}_t-\mathbf{x}_f)^{\top}\mathbf{Q}_t (\mathbf{x}_t-\mathbf{x}_f)+ \mathbf{u}_t^{\top}\mathbf{R}_t \mathbf{u}_t \big)\\
& \text{s.t.}
& & \!\!\!\!\mathbf{x}_{t+1} = \mathbf{A}_t \mathbf{x}_t + \mathbf{B}_t \mathbf{u}_t, \; i = 1, \ldots, T\\
&&& \!\!\!\!\mathbf{x}_0 = \mathbf{x}_\text{init}
\end{aligned}
\end{equation}

\end{document}